\pdfoutput=1
\documentclass[10pt,twoside]{article}
\usepackage[utf8]{inputenc}
\usepackage{amsmath,amssymb,graphicx,booktabs,multirow,textcomp}
\usepackage[letterpaper,asymmetric,textwidth=12.2cm,textheight=19.3cm,left=134.765bp,top=115.831bp,headheight=12pt,headsep=12.7pt]{geometry}
\usepackage{fancyhdr,titlesec,caption}
\usepackage[nosort]{cite}

\usepackage[hidelinks]{hyperref}
\hypersetup{pdftitle={MATE4D: Matrix-Guided Editable 4D Generation from a Single Image},pdfauthor={Xiaotian Chen and Dongfu Yin}}
\titleformat{\section}{\normalfont\large\bfseries}{\thesection}{1em}{}
\titleformat{\subsection}{\normalfont\normalsize\bfseries}{\thesubsection}{1em}{}
\titlespacing*{\section}{0pt}{15.5pt plus 3pt minus 3pt}{9.5pt}
\titlespacing*{\subsection}{0pt}{15.5pt plus 3pt minus 3pt}{5.5pt}
\newcommand{\SourcePageBreak}{\newpage}
\makeatletter
\renewcommand{\@biblabel}[1]{#1.}
\renewenvironment{thebibliography}[1]
 {\section*{References}\vspace{3.1pt}\small
  \list{\@biblabel{\@arabic\c@enumiv}}{\settowidth\labelwidth{\@biblabel{#1}}\leftmargin\labelwidth\advance\leftmargin\labelsep
  \usecounter{enumiv}\let\p@enumiv\@empty\renewcommand\theenumiv{\@arabic\c@enumiv}\setlength{\itemsep}{-0.6pt}\setlength{\parsep}{0pt}\setlength{\partopsep}{0pt}\setlength{\topsep}{0pt}}
  \sloppy\clubpenalty4000\widowpenalty4000}
 {\endlist}
\makeatother
\begin{document}
\thispagestyle{empty}
\begingroup
\centering
\vspace*{-18pt}
{\Large\bfseries MATE4D: Matrix-Guided Editable 4D\\
Generation from a Single Image\par}
\vspace{22.5pt}
{Xiaotian Chen\textsuperscript{1,2} and Dongfu Yin\textsuperscript{1$\star$}\par}
\vspace{10pt}
{\small\textsuperscript{1} Guangdong Laboratory of Artificial Intelligence and Digital Economy (SZ)\\
\textsuperscript{2} Shenzhen University\\
\texttt{chenxiaotian,yindongfu@gml.ac.cn}\par}
\endgroup
\vspace{27pt}
\begingroup\small
\setlength{\leftskip}{1cm}\setlength{\rightskip}{1cm}
\noindent
\textbf{Abstract.} Generative models have rapidly pushed content creation beyond 2D imagery toward dynamic 3D and 4D scene synthesis. Yet producing realistic and temporally stable 4D content from a single image is still difficult because one view provides limited structural cues and weak motion evidence. We introduce MATE4D, a framework that converts one input image into editable dynamic 4D content. Our method constructs a spatio-temporal multi-view image matrix with text-guided background manipulation, delivering coherent supervision over viewpoint, appearance, and motion. These synthesized observations are used to optimize 3D Gaussian primitives, which are then animated through a lightweight deformation module to form a 4D representation. The resulting scenes preserve geometry more faithfully, maintain smoother temporal behavior, and keep background edits more consistent, reducing context ambiguity and motion artifacts. Experiments on Objaverse-XL and Diffusion4D show that MATE4D outperforms strong baselines in visual quality, efficiency, and controllability, supporting practical AR/VR content creation.

\vspace{15.8pt}
\par\endgroup
\vspace{15pt}
\section{Introduction}
With the rapid progress of generative modeling, content creation has evolved from 2D images \cite{ref1} to videos \cite{ref2,ref3,ref4} and further to 3D scene generation \cite{ref5,ref6,ref7,ref8}. Building upon these advances, recent studies have explored dynamic 4D content generation \cite{ref12,ref13,ref11}, typically represented using dynamic Neural Radiance Fields (NeRFs) \cite{ref14}. Methods such as Hexplane \cite{ref15}, CONSISTENT4D \cite{ref16}, and MAV3D \cite{ref12} combine diffusion models with dynamic NeRF formulations to generate temporally varying 3D scenes from text or video inputs.

In parallel, single-image-based multi-view image generation has gained increasing attention due to its efficiency and practicality. Approaches such as

\renewcommand{\thefootnote}{$\star$}\footnotetext[1]{Corresponding author:yindongfu@gml.ac.cn}
\SourcePageBreak
\begin{center}\vspace{-\topsep}
\includegraphics[width=\linewidth]{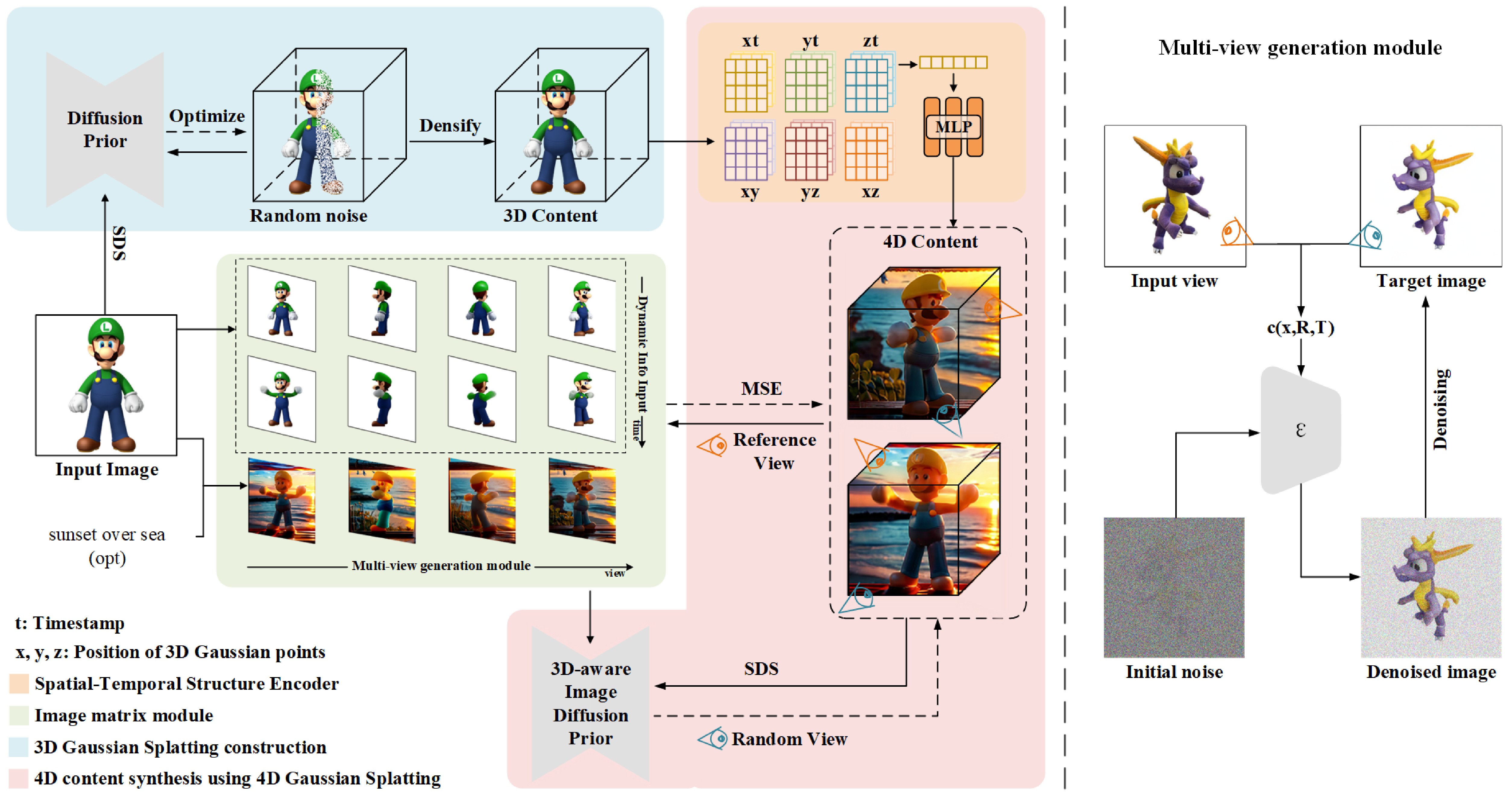}
\captionsetup{skip=13.3pt}\captionof{figure}{\textbf{The overall pipeline of MATE4D.} (Left) From a single input image, MATE4D generates a spatio-temporal multi-view image matrix that supervises 3D Gaussian Splatting and a deformation field to produce 4D content, while a text-guided module enables background editing. (Right) A fine-tuned diffusion-based generator synthesizes novel-view images for each frame, enriching the matrix with temporal and viewpoint diversity.}
\label{fig:1}
\end{center}\vspace{15pt}
\noindent Zero123 \cite{ref21}, Zero123++ \cite{ref22}, and SyncDreamer \cite{ref24} demonstrate the feasibility of synthesizing consistent novel views from a single image, providing effective supervision for downstream 3D reconstruction. On the representation side, 3D content generation has benefited from point-based, implicit, and Gaussian-based formulations. In particular, 3D Gaussian Splatting (3D GS) \cite{ref20} enables high-quality real-time rendering, and has recently been extended to dynamic scenarios through 4D Gaussian Splatting (4D GS) \cite{ref18,ref28,ref29}, offering an efficient representation for dynamic scenes with reduced optimization cost.

Despite these advances, generating high-fidelity dynamic 4D content remains challenging. Diffusion- and NeRF-based 4D methods often suffer from high computational cost, long optimization time, and limited visual sharpness. Single-image multi-view generation methods struggle to maintain geometric consistency across viewpoints, frequently leading to structural distortions, texture inconsistencies, and missing background information. These issues are further amplified when such generated views are used to supervise 3D or 4D reconstruction.

Although 4D Gaussian Splatting significantly improves rendering efficiency, existing approaches typically rely on multi-view or video inputs. When applied to single-image scenarios, they encounter fundamental limitations, including inaccurate camera estimation, insufficient background Gaussian initialization, and motion-induced surface tearing during deformation. Moreover, while background editing and lighting control have been actively studied in 2D and static 3D domains---through methods such as ControlNet \cite{ref30}, IC-Light \cite{ref31},

\SourcePageBreak
\begin{center}\vspace{-\topsep}
\includegraphics[width=\linewidth]{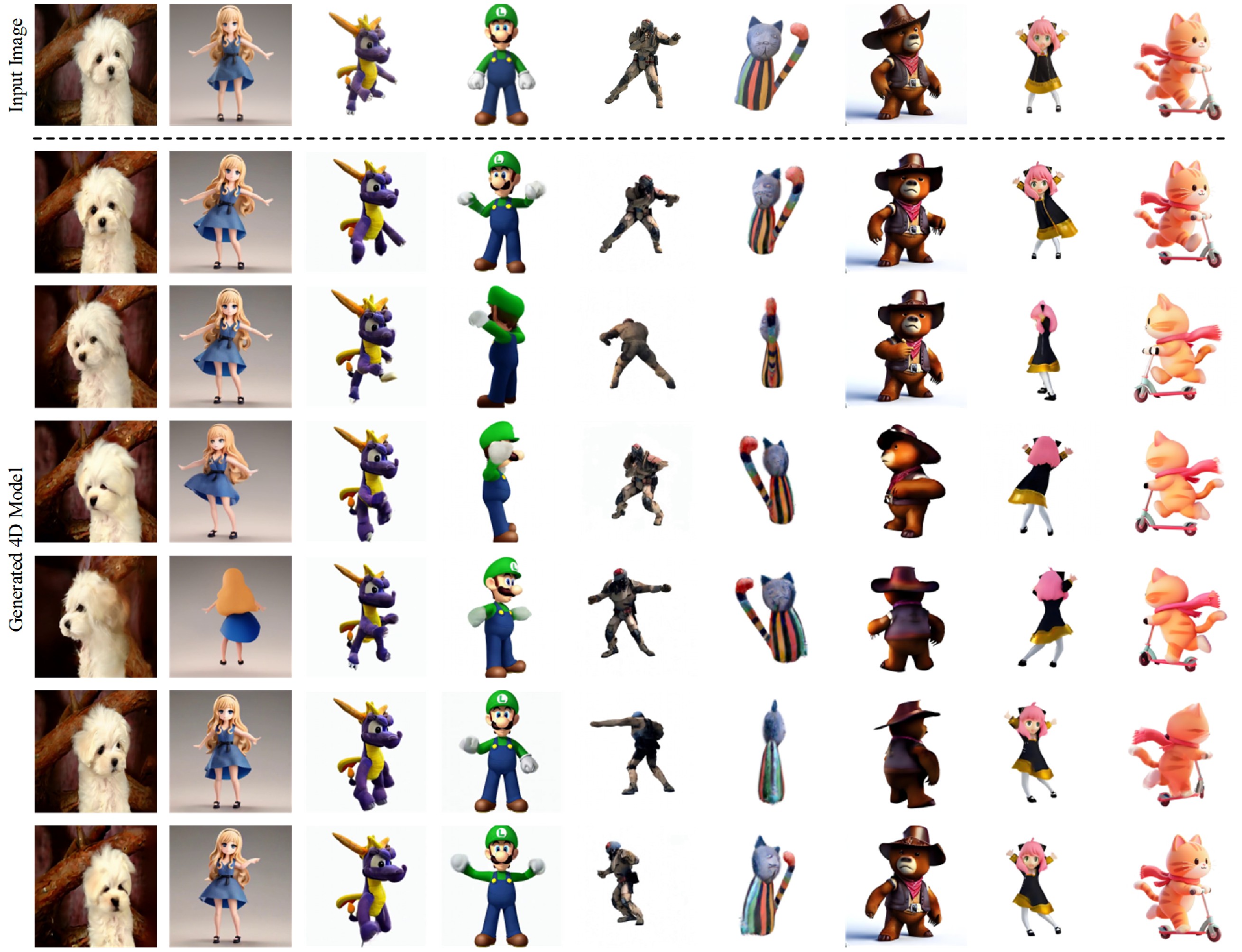}
\captionsetup{skip=13.3pt}\captionof{figure}{\textbf{Visual results.} From a single input image, MATE4D produces geometrically consistent and temporally coherent dynamic scenes. The visualized results across different timestamps and viewpoints highlight our method’s ability to preserve fine structural details and ensure smooth motion continuity without geometric distortion.}
\label{fig:2}
\end{center}\vspace{15pt}
\noindent GaussianEditor \cite{ref33}, and Scaffold-GS \cite{ref34}---such capabilities remain largely unexplored in dynamic 4D generation.

To address these limitations, we propose a novel 4D generation framework that directly generates dynamic 4D content from a single image, using 4D Gaussian Splatting as the core representation. Our method incorporates multi-view image matrix generation to enhance geometric completeness, and further introduces text and lighting supervision inspired by IC-Light to guide dynamic appearance modeling. This design effectively improves the clarity of generated 4D content, enriches motion diversity, and enables editable backgrounds, advancing practical and controllable 4D content generation from minimal input.

Our main contributions are summarized as follows:

1. We propose MATE4D, a novel single-image-to-4D generation framework that uniquely enables the creation of background-editable 4D content via text prompts.

2. We design three key components: an image matrix module that synthesizes temporally coherent multi-view sequences, a 4D Gaussian Splatting representa-

\SourcePageBreak\noindent

tion with a lightweight deformation field for dynamic modeling, and a text-guided background editing module for realistic environmental control.

\vspace{8pt}
3. We conduct extensive experiments on large-scale benchmarks, including Objaverse-XL and Diffusion4D, and demonstrate that our method achieves state-of-the-art performance in spatial fidelity, temporal coherence, and editing flexibility, surpassing existing 4D generation approaches.

\vspace{56pt}
\begin{center}\vspace{-\topsep}
\includegraphics[width=\linewidth]{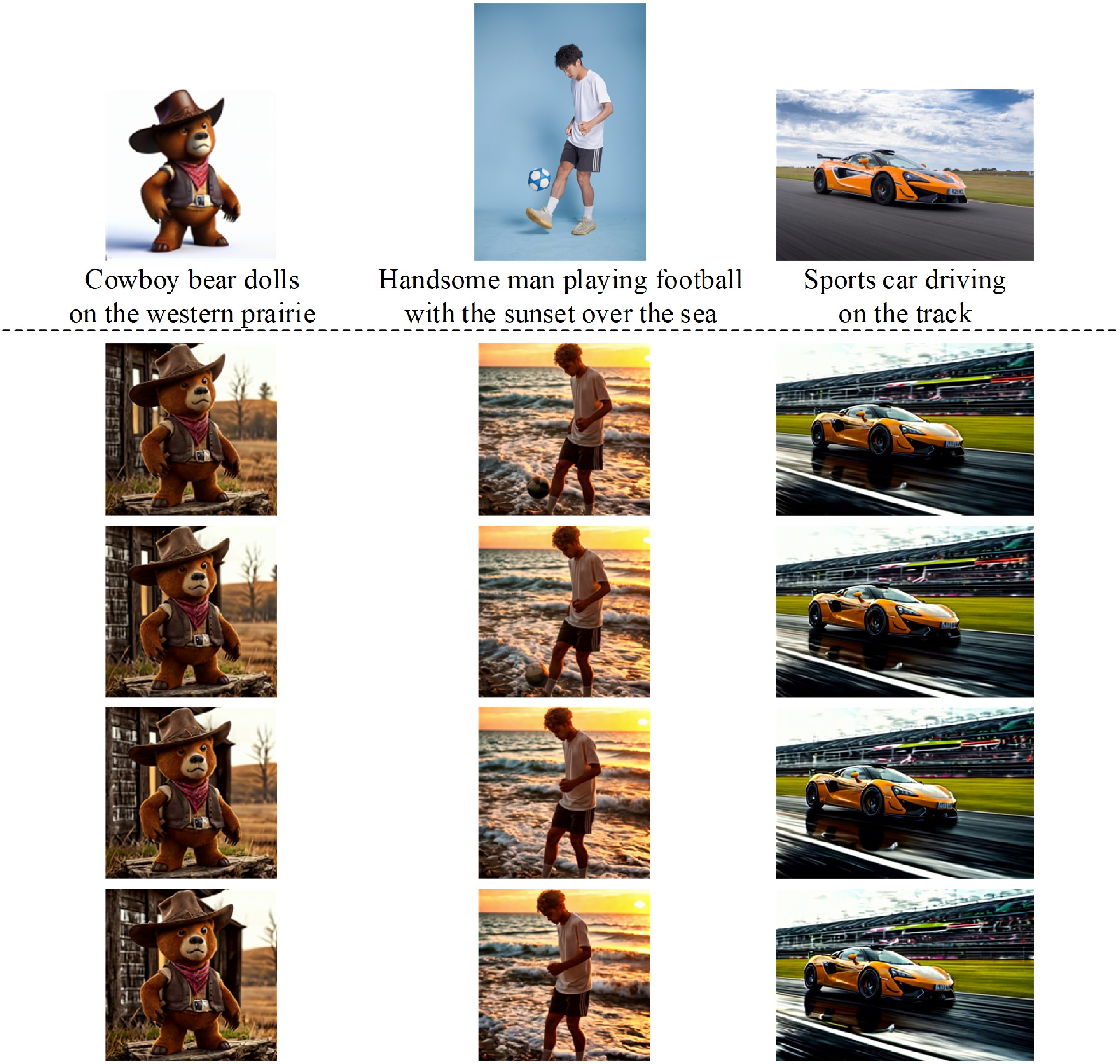}
\captionsetup{skip=13.3pt}\captionof{figure}{Dynamic 4D frames from a single edited view via text-guided background editing, with consistent motion and viewpoints, showing 4D foreground-background generation.}
\label{fig:3}
\end{center}\vspace{15pt}
\SourcePageBreak
\section{Related Work}
\subsection{Single-image multi-view generation}
Single-image novel-view synthesis has rapidly advanced with the development of diffusion priors. Zero-1-to-3 conditions a large diffusion model on relative camera poses to synthesize new viewpoints from a single image \cite{ref21}. Zero123++ improves cross-view consistency through joint generation of tiled multi-view outputs \cite{ref22}. SyncDreamer and Wonder3D further strengthen 3D awareness by modeling multi-view correlations and auxiliary geometric cues \cite{ref24,ref5}. Although these methods provide effective supervision for static 3D reconstruction, they do not explicitly address the joint temporal-view consistency required for editable 4D generation.

\subsection{4D content generation}
Early 4D generation methods are largely based on dynamic NeRF formulations, where time is introduced into implicit radiance field models to represent deforming scenes. More recent work distills video diffusion priors into dynamic 3D representations, enabling text- or image-driven 4D generation \cite{ref12,ref16,ref17}. However, such pipelines often suffer from high optimization cost, unstable geometry, and limited rendering efficiency. Gaussian-based representations offer a more practical alternative. 3D Gaussian Splatting achieves high-quality real-time rendering with explicit point primitives \cite{ref20}, and 4D Gaussian variants extend this idea to dynamic scenes by coupling a canonical point cloud with time-dependent deformations \cite{ref18,ref29,ref32}. Still, existing methods rarely consider editable environments as part of the 4D learning objective.

\subsection{Editable background and lighting control}
Text-guided control and relighting have been extensively studied in 2D generation. ControlNet injects structural conditions into diffusion backbones for controllable image generation \cite{ref30}. Recent relighting methods, such as IC-Light, emphasize consistent illumination transfer while preserving subject appearance \cite{ref31}. In static 3D editing, Gaussian-based scene editing frameworks show that explicit representations are well suited to controllable appearance manipulation \cite{ref33}. Our work differs from these settings by incorporating editable background and lighting control directly into dynamic 4D generation from a single image.

\section{Method}
The proposed MATE4D framework generates spatio-temporally coherent 4D content from a single input image. As illustrated in Fig~\ref{fig:1}, it consists of three components: an \textbf{image matrix module} for constructing coherent multi-view sequences, a \textbf{4D content synthesis} stage based on Gaussian Splatting, and a \textbf{background editing module} that enables flexible control of illumination and environment.

\SourcePageBreak
\begin{center}\vspace{-\topsep}
\captionof{table}{Quantitative comparison of visual quality with state-of-the-art methods on \textbf{Objaverse-XL dataset.} \textbf{MATE4D}$^{\dagger}$ represents our model evaluated with foreground masks for a fair comparison with baselines. \textbf{MATE4D} denotes evaluation on the full image including text-guided edited backgrounds.}
\label{tab:1}
\small
\begin{tabular*}{\linewidth}{@{\extracolsep{\fill}}l|ccccc@{}}
\toprule
Methods & Background & CLIP-I$\uparrow$ & LPIPS$\downarrow$ & SSIM$\uparrow$ & PSNR$\uparrow$ \\
\midrule
Animate124 \cite{ref17} & & 0.854 & 0.132 & 0.812 & 33.841 \\
DG4D \cite{ref28} & $\checkmark$ & 0.923 & 0.145 & 0.764 & 34.054 \\
V4D \cite{ref41} & $\checkmark$ & 0.889 & 0.131 & 0.89 & 33.726 \\
Consistent4D \cite{ref16} & & 0.921 & 0.165 & 0.841 & 34.102 \\
EG4D \cite{ref38} & & 0.932 & 0.133 & 0.857 & 34.211 \\
Diffusion4D \cite{ref32} & & 0.905 & 0.134 & 0.868 & 34.257 \\
STAG4D \cite{ref43} & & 0.918 & 0.126 & 0.871 & 35.077 \\
Efficient4D \cite{ref44} & & 0.912 & 0.146 & 0.875 & 34.289 \\
4DGen \cite{ref45} & & 0.927 & \textbf{0.124} & 0.879 & 34.362 \\
Gaussian-Flow \cite{ref46} & $\checkmark$ & 0.949 & 0.146 & 0.904 & 34.475 \\
SV4D \cite{ref29} & $\checkmark$ & 0.913 & 0.129 & \textbf{0.952} & 34.583 \\
\midrule
MATE4D$^{\dagger}$ (Ours) & & \textbf{0.972} & 0.214 & 0.884 & 36.154 \\
MATE4D(Ours) & $\checkmark$ & 0.961 & 0.152 & 0.936 & \textbf{38.762} \\
\bottomrule
\end{tabular*}
\end{center}\vspace{0pt}

\vspace{3.1pt}
\subsection{Image Matrix Module}
\vspace{6.8pt}
Given a single input image $I_0$, we first transform it into a dynamic video sequence $\{I_t\}_{t=1}^T$ to introduce temporal information. Since each frame $I_t$ contains only a single viewpoint, we design a view-conditioned generation mechanism that synthesizes novel views from relative camera parameters $(\Delta\theta,\Delta\phi,\Delta r)$. Specifically, we fine-tune a diffusion-based multi-view generation model (Fig~\ref{fig:1}, right) to produce consistent novel viewpoints for each frame, conditioned on the input image and relative camera transformation. This ensures that the resulting image matrix preserves both temporal dynamics and viewpoint diversity. Formally, the training objective is defined as:

\begingroup
\setlength{\abovedisplayskip}{13.5pt}
\setlength{\belowdisplayskip}{13.5pt}
\begin{equation}
\min_{\theta}\;\mathbb E_{z\sim\mathcal E(x),t,\epsilon\sim\mathcal N(0,1)}
\left\|\epsilon-\epsilon_\theta(z_t,t,c(x,R,T))\right\|_2^2,
\label{eq:diffusion}
\end{equation}
\endgroup
\noindent where $x$ denotes the input frame, $c(x,R,T)$ is the embedding that fuses image features with relative viewpoint, $\epsilon_\theta$ is a U-Net denoiser, and $z_t$ is the noisy latent at timestep $t$.

By applying this procedure across $T$ frames and $V$ viewpoints, we obtain an image matrix $\mathcal M$ encoding both temporal dynamics and multi-view diversity. To maintain spatio-temporal coherence, we further enforce perceptual alignment between generated frames and their inputs, rather than treating each viewpoint independently.

\SourcePageBreak
\begin{center}\vspace{-\topsep}
\captionof{table}{Quantitative comparison of visual consistency with state-of-the-art methods on \textbf{Objaverse-XL dataset.} \textbf{MATE4D}$^{\dagger}$ represents our model evaluated with foreground masks for a fair comparison with baselines. \textbf{MATE4D} denotes evaluation on the full image including text-guided edited backgrounds.}
\label{tab:2}
\small
\begin{tabular*}{\linewidth}{@{\extracolsep{\fill}}l|ccccc@{}}
\toprule
Methods & Background & FVD-F$\downarrow$ & FVD-V$\downarrow$ & FVD-Diag$\downarrow$ & FVD4D$\downarrow$ \\
\midrule
Animate124 \cite{ref17} & & 725.947 & 392.617 & 553.847 & 559.347 \\
DG4D \cite{ref28} & $\checkmark$ & 697.843 & 450.585 & 615.681 & 638.150 \\
V4D \cite{ref41} & $\checkmark$ & 792.536 & 417.853 & 589.326 & 602.419 \\
Consistent4D \cite{ref16} & & 1133.931 & 488.382 & 741.526 & 871.950 \\
EG4D \cite{ref38} & & 814.229 & 432.941 & 612.719 & 641.238 \\
Diffusion4D \cite{ref32} & & 836.742 & 440.328 & 624.538 & 655.842 \\
STAG4D \cite{ref43} & & 861.881 & 469.074 & 636.832 & 546.560 \\
Efficient4D \cite{ref44} & & 848.913 & 447.512 & 635.217 & 678.527 \\
4DGen \cite{ref45} & & 865.327 & 459.837 & 525.656 & 702.316 \\
Gaussian-Flow \cite{ref46} & $\checkmark$ & 956.567 & 490.777 & 702.736 & 762.753 \\
SV4D \cite{ref29} & $\checkmark$ & 677.686 & \textbf{368.538} & 648.932 & 614.350 \\
\midrule
MATE4D$^{\dagger}$ (Ours) & & 258.912 & 412.342 & 232.973 & 192.423 \\
MATE4D(Ours) & $\checkmark$ & \textbf{236.338} & 395.536 & \textbf{214.359} & \textbf{175.828} \\
\bottomrule
\end{tabular*}
\end{center}\vspace{0pt}

\subsection{4D Content Synthesis}
Based on the constructed image matrix $\mathcal M$, we initialize a 3D Gaussian Splatting representation $\mathcal G_0=\{g_i\}_{i=1}^N$ and extend it into a dynamic 4D representation via a spatio-temporal deformation field. Each Gaussian $g_i$ is parameterized by its $(x,y,z)$ position, covariance, opacity, and spherical harmonics coefficients. The deformation field $D_\phi$ predicts time-dependent transformations $g_i^t=D_\phi(g_i,t)$, resulting in a set of 4D Gaussians $\mathcal G_{4D}$.

To optimize this representation, we render each viewpoint $\hat I_{t,v}$ using differentiable splatting and minimize its discrepancy with the image matrix:

\begin{equation}
\mathcal L_{Ref}=\frac{1}{TV}\sum_{t=1}^T\sum_{v=1}^V
\left\|\hat I_{t,v}-I_{t,v}\right\|_2^2.
\label{eq:reference}
\end{equation}
In addition, to ensure generalization to unseen viewpoints and recovery of occluded structures, we incorporate a score distillation sampling (SDS) loss with a pretrained 2D diffusion prior:

\begin{equation}
\nabla_\Theta\mathcal L_{SDS}=\mathbb E_{t,p,\epsilon}\left[
\omega(t)\left(\epsilon_\phi(\hat I;t,\hat I^{r},\triangle p)-\epsilon\right)
\frac{\partial\hat I}{\partial\Theta}\right],
\label{eq:sds}
\end{equation}
\noindent where $\triangle p$ denotes the relative pose change, $\omega(t)$ is the timestep-dependent weight, and $\epsilon_\phi$ is the noise predicted by the diffusion prior. The joint objective $\mathcal L_{Ref}+\lambda\mathcal L_{SDS}$ thus balances data fidelity with prior-guided plausibility.

\SourcePageBreak
\begin{center}\vspace{-\topsep}
\includegraphics[width=\linewidth]{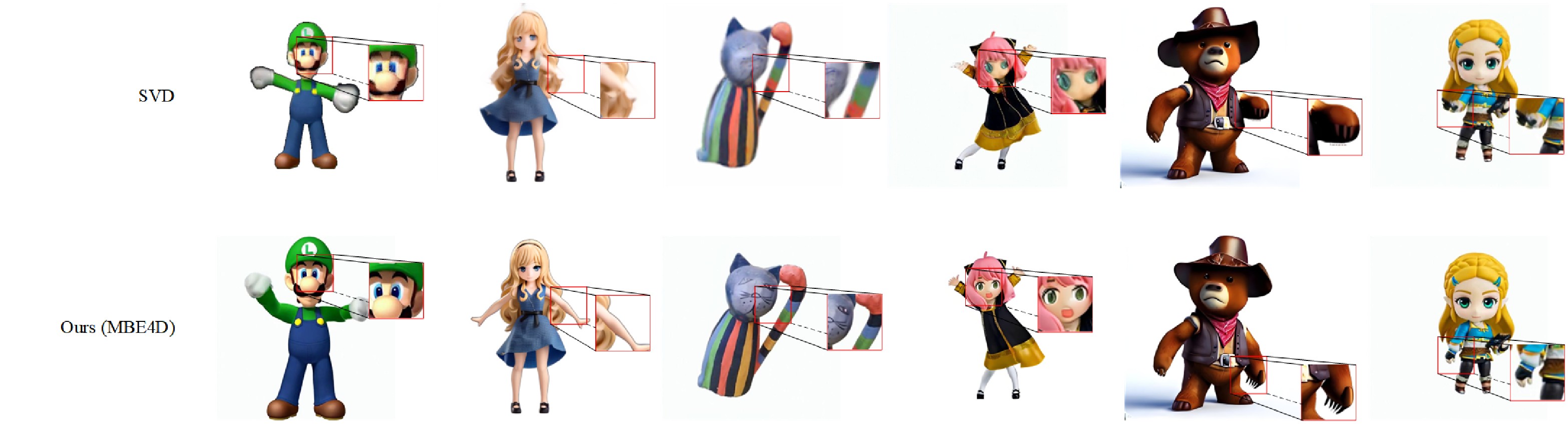}
\captionsetup{skip=13.3pt}\captionof{figure}{\textbf{Detail enlarge comparison diagram.} This figure illustrates the comparative performance of the baseline method (SVD) and our proposed method (MATE4D) in terms of detail recovery and boundary treatment.}
\label{fig:4}
\end{center}\vspace{15pt}
\begin{center}\vspace{-\topsep}
\captionof{table}{\textbf{Regarding the quantitative results of MATE4D on the Diffusion4D dataset.} The lower the value, the better the performance.}
\label{tab:3}
\small
\begin{tabular*}{\linewidth}{@{\extracolsep{\fill}}lccc@{}}
\toprule
Methods & FVD-F$\downarrow$ & FVD-Diag$\downarrow$ & FVD4D$\downarrow$ \\
\midrule
DG4D \cite{ref28} & 487.207 & 702.853 & 731.330 \\
V4D \cite{ref41} & 489.728 & 688.939 & 643.227 \\
Consistent4D \cite{ref16} & 481.891 & 672.648 & 752.762 \\
EG4D \cite{ref38} & 463.518 & 674.495 & 728.486 \\
Diffusion4D \cite{ref32} & 437.921 & 596.207 & 598.456 \\
STAG4D \cite{ref43} & 456.227 & 599.936 & 536.330 \\
Efficient4D \cite{ref44} & 415.793 & 628.828 & 672.768 \\
4DGen \cite{ref45} & 408.087 & 593.873 & 539.441 \\
Gaussian-Flow \cite{ref46} & 370.532 & 555.611 & 618.817 \\
SV4D \cite{ref29} & 397.279 & 543.332 & 598.021 \\
\midrule
MATE4D(ours) & \textbf{273.924} & \textbf{457.342} & \textbf{397.346} \\
\bottomrule
\end{tabular*}
\end{center}\vspace{0pt}

\subsection{Background Editing}
To enable editing of background and illumination, we introduce a text-guided background editing module. Given an input image $I_0$ and a textual description $T$, we generate a background-enhanced image:

\begin{equation}
I_0^{bg}=f_{bg}(I_0,T),
\label{eq:background}
\end{equation}
\noindent where $f_{bg}$ modifies the environment while preserving object integrity.

The edited image $I_0^{bg}$ is processed by the same image matrix module, producing a background-aware matrix $\mathcal M_{bg}$. To disentangle background from object geometry, we initialize a background-free Gaussian point cloud and jointly optimize it with $\mathcal M_{bg}$. This design enabling background-editable 4D synthesis under varying illumination and environmental conditions.

\SourcePageBreak
\titlespacing*{\subsection}{0pt}{20pt}{11.5pt}
\section{Experiments}
\subsection{Datasets and Evaluation Metrics}
\textbf{Objaverse-XL.} We evaluate our method on the Objaverse-XL dataset \cite{ref35}, a large-scale repository of over 10 million 3D objects spanning diverse categories. For 4D generation tasks, we utilize rendered multi-view sequences of these objects, enabling assessment of our model’s ability to synthesize temporally consistent dynamic scenes from complex geometries.

\textbf{Diffusion4D.} To further validate our approach, we use the Diffusion4D dataset \cite{ref32}, which provides dynamic 4D sequences generated via diffusion-based models. It contains a wide range of objects and motions, capturing realistic appearance variations and temporal dynamics. This dataset serves as a challenging benchmark for evaluating both spatial fidelity and temporal coherence.

\textbf{Evaluation Metrics.} To measure visual quality, we adopt CLIP-I, PSNR, SSIM, and LPIPS. To evaluate motion consistency, we employ a suite of FVD metrics, including FVD-F, FVD-V, FVD-Diag, and FVD4D, which capture temporal fidelity, viewpoint consistency, and deformation accuracy in generated 4D content.

\vspace{3.3pt}
\subsection{Implementation Details}
\vspace{3.1pt}
In the multi-view generation module, we fix the rotation along the y-axis and vary the x-axis rotation with offsets $\Delta x=[0,10,20,10,0,-10,-20,-10,0]$ for the second stage. To introduce dynamic information, we adopt SVD as the temporal modeling backbone. For background editing, images are rendered at a resolution of $512\times512$. During 4D Gaussian Splatting optimization, we supervise reconstruction using batches of 9 images sampled from the multi-view image matrix.

\vspace{20pt}
\setlength{\tabcolsep}{3.78pt}
\begin{center}\vspace{-\topsep}
\captionof{table}{\textbf{Performance comparison between RGBA and RGB image formats.} Quantitative comparison of MB4D and SV4D across multiple datasets using RGBA and RGB formats. Performance is evaluated via CLIP-I and PSNR metrics.}
\label{tab:4}
\small
\resizebox{\linewidth}{!}{%
\begin{tabular}{@{}cc|ccc|ccc@{}}
\toprule
\multicolumn{2}{c|}{Data Format} & \multicolumn{3}{c|}{RGBA} & \multicolumn{3}{c}{RGB} \\
\midrule
\multicolumn{2}{c|}{Data Source} & Unsplash & Dalle & Objaverse & Unsplash & Dalle & Objaverse \\
\midrule
\multirow{2}{*}{SV4D} & CLIP-I & 0.967 & 0.965 & 0.958 & 0.941 & 0.954 & 0.915 \\
& PSNR & 35.97 & 34.12 & 33.68 & 28.12 & 32.88 & 27.59 \\
\midrule
\multirow{2}{*}{\shortstack{MATE4D\\(Ours)}} & CLIP-I & \textbf{0.979}$\uparrow$ & 0.969 & 0.941 & 0.913 & 0.973 & 0.921 \\
& PSNR & 34.13 & 35.94 & \textbf{36.44}$\uparrow$ & 27.56 & 32.35 & 35.20 \\
\bottomrule
\end{tabular}}
\end{center}\vspace{0pt}

\setlength{\tabcolsep}{3.5pt}
\SourcePageBreak
\titlespacing*{\subsection}{0pt}{15.5pt plus 3pt minus 3pt}{5.5pt}
\begin{center}\vspace{-\topsep}
\captionof{table}{Ablation study of IMM, BEM, and DII.}
\label{tab:5}
\small
\begin{tabular*}{\linewidth}{@{\extracolsep{\fill}}lcccccc@{}}
\toprule
 & IMM & BEM & DII & CLIP-I$\uparrow$ & PSNR$\uparrow$ & FVD4D$\downarrow$ \\
\midrule
(a) & & & & 0.859 & 30.54 & - \\
(b) & & $\checkmark$ & & 0.949 & 35.734 & - \\
(c) & $\checkmark$ & & $\checkmark$ & 0.957 & 36.245 & 213.653 \\
(d) & & $\checkmark$ & $\checkmark$ & 0.861 & 30.441 & 421.562 \\
(e) & $\checkmark$ & $\checkmark$ & $\checkmark$ & \textbf{0.961} & \textbf{38.762} & \textbf{175.828} \\
\bottomrule
\end{tabular*}
\end{center}\vspace{0pt}

\subsection{Main Results}
As shown in Table~\ref{tab:1} and Table~\ref{tab:2}, MATE4D achieves state-of-the-art performance across most metrics, specifically reaching the highest CLIP-I (0.961), PSNR (38.762), and SSIM (0.936). These results, along with the lowest FVD scores, verify its superior ability to preserve spatial fidelity and temporal coherence. To further evaluate our background editing capability, we report results in two modes: MATE4D (evaluated on unedited images for fair comparison with baselines) and MATE4D$^{\dagger}$ (full-image evaluation after background editing). While MATE4D$^{\dagger}$ shows an expected decrease in pixel-wise metrics (PSNR/SSIM) due to the divergence between the synthesized and original background pixels, its CLIP-I score increases to 0.972, and the FVD metrics remain competitive. This gap demonstrates that our Background Editing Module (BEM) significantly enhances global semantic coherence and temporal stability while maintaining competitive motion quality, effectively balancing high-fidelity reconstruction with versatile editability.

At present, mainstream image formats include RGBA and RGA. We evaluated our method across different formats on multiple datasets using CLIP-I and PSNR, comparing against SV4D\cite{ref29} as the baseline. As shown in Table~\ref{tab:4}, our method consistently achieves better performance. The generation quality improves notably with RGBA inputs, likely because the alpha channel provides useful semantic cues for 4D reconstruction. Performance further increases with grayscale inputs, indicating that their concise and robust representation benefits downstream tasks and future research.

Beyond quantitative results, qualitative comparisons further validate the robustness of our framework. As illustrated in Fig~\ref{fig:2} and Fig~\ref{fig:3}, MATE4D generates realistic and temporally coherent 4D sequences from a single image, faithfully preserving fine-grained textures such as clothing folds and fur details, while maintaining stable geometry under challenging rotations. The smooth transitions across viewpoints and timestamps demonstrate that the learned 4D Gaussian representation effectively mitigates motion discontinuities and geometric distortions, issues frequently observed in prior approaches.

As shown in Fig~\ref{fig:4}, we further compare our MATE4D with Stable Video Diffusion (SVD) \cite{ref2}. While SVD suffers from blurred high-frequency details, boundary noise, and jagged edges, our method produces sharper contours, suppresses

\SourcePageBreak
\begin{center}\vspace{-\topsep}
\captionof{table}{Quantitative ablation study of loss components on the \textbf{Diffusion4D dataset.}}
\label{tab:6}
\small
\begin{tabular}{@{}lcccc@{}}
\toprule
Method & PSNR$\uparrow$ & SSIM$\uparrow$ & CLIP-I$\uparrow$ & FVD4D$\downarrow$ \\
\midrule
w/o $\mathcal L_{SDS}$ ($\mathcal L_{MSE}$ Only) & 39.12 & 0.941 & 0.912 & 245.5 \\
w/o $\mathcal L_{MSE}$ ($\mathcal L_{SDS}$ Only) & 24.35 & 0.721 & 0.885 & 310.2 \\
Full Model ($\mathcal L_{MSE}+\mathcal L_{SDS}$) & \textbf{38.76} & \textbf{0.936} & \textbf{0.961} & \textbf{175.8} \\
\bottomrule
\end{tabular}
\end{center}\vspace{0pt}

\vspace{13.3pt}
\noindent artifacts more effectively, and delivers smoother, more natural transitions. This results in significantly improved visual fidelity and structural integrity in the generated 4D content.

We further evaluate on the Diffusion4D dataset, with results presented in Table~\ref{tab:3}. MATE4D consistently achieves the lowest FVD-based scores, demonstrating strong generalization to challenging dynamic motions and diverse object categories.

\vspace{3.5pt}
\subsection{Ablation Studies}
\vspace{3.2pt}
To evaluate the contribution of each component, we conduct ablation studies on the Image Matrix Module (IMM), the Background Editing Module (BEM), and the Dynamic Info Input (DII). Quantitative results are summarized in Table~\ref{tab:5}. The base configuration without any proposed modules (a) fails to generate reliable 4D content, yielding the lowest CLIP-I (0.859) and PSNR (30.54). Introducing IMM alone (b) significantly enhances spatial fidelity (CLIP-I 0.949, PSNR 35.734), underscoring its essential role in multi-view consistency. Adding DII (c) further improves semantic and pixel-level accuracy while markedly reducing FVD4D to 213.653, which validates the importance of temporal dynamics for motion coherence. Conversely, the configuration omitting IMM (d) performs poorly (PSNR 30.441, FVD4D 421.562), highlighting that IMM is indispensable for stable geometry. Our full model (e) achieves the best performance across all metrics (CLIP-I 0.962, PSNR 38.762, FVD4D 175.828). In this setting, BEM complements IMM and DII by further enhancing visual quality and background consistency. These results demonstrate that the three modules are mutually beneficial and essential for high-fidelity, temporally coherent 4D generation, confirming the effectiveness of our framework.

To further verify the effectiveness of the MATE4D optimization objective, we conducted a detailed ablation study on the reconstruction loss $\mathcal L_{MSE}$ and the score distillation sampling loss $\mathcal L_{SDS}$. As shown in Table~\ref{tab:6}, the experimental results indicate that these two components play complementary roles in balancing the generation quality and geometric fidelity. Specifically, when the reconstruction loss $\mathcal L_{MSE}$ is removed and only SDS is relied upon for guidance, the PSNR drops sharply from 38.76 dB to 24.35 dB, which strongly demonstrates that explicit pixel supervision based on the image matrix is crucial for maintaining high fidelity of the generated assets with the input images and preventing texture

\SourcePageBreak
\noindent drift. On the other hand, although the PSNR in the reference view slightly increases after removing $\mathcal L_{SDS}$ (due to overfitting of the model in the limited view), the CLIP-I and FVD4D metrics, which reflect the overall generation quality and spatiotemporal coherence, deteriorate significantly.

\section{Conclusion}
\vspace{6.1pt}
We introduced MATE4D, a framework for generating dynamic 4D content from a single image. By integrating a spatio-temporal image matrix, a Gaussian-based deformation representation, and a background editing module, our method achieves geometric realism, smooth temporal dynamics, and controllable environmental conditions. Comprehensive evaluations on large-scale benchmarks demonstrate that MATE4D delivers state-of-the-art performance in both spatial fidelity and temporal coherence, while providing enhanced flexibility for immersive AR/VR applications. This work takes a step toward practical and editable 4D generation from minimal inputs. Future work will focus on further reducing computational overhead, while enhancing generalization to diverse application scenarios and enriching the diversity of dynamic motions.

\vspace{3.1pt}
\section{Acknowledgments}
\vspace{3.2pt}
This work was supported by the Shenzhen Science and Technology Program under Grant KJZD20240903104400001 and was financially supported by the Research Task Assignment Project from Guangdong Laboratory of Artificial Intelligence and Digital Economy (SZ), under Grant No. GML-26420004.

\end{document}